\documentclass[a4paper,fleqn]{cas-dc}
\usepackage[numbers]{natbib}
\usepackage{tikz,lipsum,lmodern}
\usepackage[most]{tcolorbox}
\usepackage{enumitem} 
\usepackage{siunitx}
\usepackage{listings}     
\usepackage{colortbl}  
\usepackage{graphicx}   
\usepackage{array}      
\usepackage{graphicx}
\usepackage{subcaption}
\usepackage{amsmath,amssymb}

\usepackage{multirow}   
\usepackage{adjustbox}  
\usepackage{caption}
\usepackage{hhline}    
\usepackage{graphicx}     
\usepackage{array}        
\usepackage{adjustbox}    
\usepackage{caption}      
\usepackage{tabularx}     
\usepackage{natbib}
\usepackage{amsmath}
\usepackage{regexpatch}

\tcbset{
  refusal/.style={
    colback=red!20, 
    colframe=red!50, 
    boxrule=0.5mm, 
    sharp corners, 
    left=2mm, 
    right=2mm, 
    top=1mm, 
    bottom=1mm, 
  }
}

\renewcommand{\arraystretch}{1} 

\lstdefinelanguage{json}{
    basicstyle=\ttfamily\footnotesize,
    showstringspaces=false,
    breaklines=true,
    frame=single,
    literate=
     *{0}{{{\color{red}0}}}{1}
      {1}{{{\color{red}1}}}{1}
      {2}{{{\color{red}2}}}{1}
      {3}{{{\color{red}3}}}{1}
      {4}{{{\color{red}4}}}{1}
      {5}{{{\color{red}5}}}{1}
      {6}{{{\color{red}6}}}{1}
      {7}{{{\color{red}7}}}{1}
      {8}{{{\color{red}8}}}{1}
      {9}{{{\color{red}9}}}{1}
      {:}{{{\color{blue}:}}}{1}
      {,}{{{\color{blue},}}}{1}
      {"}{{{\color{green}"}}}{1}
      {\{}{{{\color{magenta}\{}}}{1}
      {\}}{{{\color{magenta}\}}}}{1}
}

\definecolor{high}{RGB}{198,239,206}   
\definecolor{medium}{RGB}{255,235,156} 
\definecolor{low}{RGB}{255,199,206}    
\definecolor{header}{RGB}{220,230,241} 
\definecolor{best}{RGB}{198,239,206}    
\definecolor{second}{RGB}{184,204,228}  
\definecolor{worst}{RGB}{255,199,206}   

\definecolor{lightred}{HTML}{FDEDEC} 
\usepackage{arydshln}

\def\tsc#1{\csdef{#1}{\textsc{\lowercase{#1}}\xspace}}
\tsc{WGM}
\tsc{QE}
\tsc{EP}
\tsc{PMS}
\tsc{BEC}
\tsc{DE}

\begin{document}
\let\WriteBookmarks\relax
\def\floatpagepagefraction{1}
\def\textpagefraction{.001}

\shorttitle{Leveraging social media news}

\shortauthors{Hambarde, K.A. et~al.}

\title [mode = title]{Human Re-ID Meets LVLMs: What can we expect?}                      

\tnotetext[1]{This work was funded by FCT/MEC through national funds and co-funded by the FEDER—PT2020 partnership agreement under the projects UIDB \newline/50008/2020 and POCI-01-0247-FEDER-033395.}


%
\author[1]{Kailash A. Hambarde}[type=editor,
                        auid=000,bioid=1,
                        orcid=0000-0003-1012-2952
                        ]

\cormark[1]


\ead{kailas.srt@gmail.com}

\credit{Conceptualization, Methodology, Software, Writing - initial draft preparation}

\affiliation[1]{organization={Department of Computer Science, University of Beira Interior },
    city={Covilhã},
    postcode={6201-001}, 
    country={Portugal}}


\author[1]{Pranita Samale}[
    orcid=0009-0009-9614-7755
]

\credit{Conceptualization, Methodology, Software, Writing - initial draft preparation}

\author[1]{Hugo Proença}[                      orcid=0000-0003-2551-8570]


\credit{Conceptualization, Methodology, Review - final draft preparation}



\cortext[cor1]{Corresponding author}
\cortext[cor2]{Principal corresponding author}



\begin{abstract}
Large vision-language models (LVLMs) have been regraded as a breakthrough advance in an astoundingly variety of tasks, from content generation to virtual assistants and multimodal search/retrieval. However, for many of these applications, the  performance of these methods has been widely criticized, particularly when compared with state-of-the-art methods and technologies in each specific domain. In this work, we  compare the performance of the leading large vision-language models in the human re-identification task, using as baseline the performance attained by state-of-the-art AI models specifically designed for this problem. We compare the results due to \textbf{ChatGPT-4o}, \textbf{Gemini-2.0-Flash}, \textbf{Claude 3.5 Sonnet}, and \textbf{Qwen-VL-Max} to a baseline ReID \textbf{PersonViT} model, using the well known  Market1501 dataset.  Our evaluation pipeline includes  the dataset curation, prompt engineering, and metric selection to assess the models' performance. Results are analyzed from many different perspectives: similarity scores, classification accuracy, and classification metrics, including precision, recall, f1 score, and area under curve (AUC). Our results confirm the strengths of LVLMs, but also their severe limitations that often lead to catastrophic answers, and should be the scope of further research.  As concluding remark, we speculate about some further research that should fuse traditional and LVLMs to combine the strengths from both families of techniques and c+achieve solid improvements in performance.  

\end{abstract}



\begin{keywords}
    Large Vision-Language Models \sep 
    Human Re-identification \sep 
    ChatGPT-4o \sep 
    Claude-3.5 \sep
    Gemini-2.0 \sep
    Qwen-VL-Max \sep
\end{keywords}

\maketitle

\section{Introduction}

Human re-identification (ReID) is a critical task in computer vision enabling the identification of individuals across multiple cameras in surveillance systems \cite{hambarde2024image}. The growing ubiquity of surveillance cameras in public spaces has created an increasing demand for accurate and efficient ReID systems. Traditionally specialized models such as PersonViT \cite{hu2024personvit} have excelled in this domain demonstrating strong performance on datasets like Market1501 \cite{zheng2015scalable} which features low resolution images and diverse surveillance conditions. However with the rise of large vision-language models (LVLMs) which have shown impressive versatility and adaptability across various computer vision tasks there is a compelling need to explore their potential for ReID task. Recent advancements in LVLMs have demonstrated significant potential in tackling a wide range of computer vision tasks including object detection \cite{zang2024contextual, han2024few}, activity and gesture recognition \cite{ji2024hargpt, bimbraw2024gpt}, and facial analysis tasks such as expression recognition, face recognition, and age estimation \cite{zhao2023prompting, aldahoul2024exploring, hassanpour2024chatgpt}, deep-fakes detection \cite{jia2024can}, iris recognition \cite{farmanifard2024chatgpt}. Inspired by these successes this study evaluates the performance of leading LVLMs including OpenAI ChatGPT-4o \cite{achiam2023gpt}, Google Gemini 2.0 \cite{google2024gemini}, Anthropic Claude 3.5 \cite{anthropic2024claude}, and Alibaba Qwen-VL Max \cite{Qwen-VL} for Human ReID tasks.

Our study employs a structured evaluation pipeline including dataset curation, prompt engineering, and robust evaluation metrics, such as similarity scores, precision, recall, f1 score, and area under curve (AUC). By comparing LVLMs to specialized ReID model on the Market1501 dataset we assess their robustness in handling realistic challenges posed by low resolution images and diverse surveillance environments.
This study addresses a key question: Can LVLMs effectively handle Human ReID tasks especially in challenging surveillance conditions? By analyzing their performance we provide insights into their strengths limitations to help for further research and development in ReID systems and LVLMs.

\section{Background}
Recent progress in LVLMs has brought together the fields of computer vision and natural language processing (NLP). These transformer-based models \cite{vaswani2017attention} process both images and text achieving strong performance on tasks such as image classification, object detection, semantic segmentation, language modeling, and question answering. By integrating visual and linguistic information LVLMs gain a more detailed understanding of how images and text relate to each other \cite{zhang2024vision}.
Several studies have illustrated the capabilities of LVLMs across various domains. For example, Shan Jia et al. \cite{jia2024can} demonstrated that these models can effectively detect deepfakes accurately identifying manipulated images. Similarly LVLMs have proven successful in facial analysis tasks such as expression recognition, face recognition, age estimation, and iris recognition by leveraging their multimodal nature to analyze facial features, expressions, and contextual cues with impressive accuracy \cite{zhao2023prompting, aldahoul2024exploring, hassanpour2024chatgpt, farmanifard2024chatgpt}.
Despite these achievements the role of LVLMs in ReID remains largely unexplored. To fill this gap we evaluate the performance of leading LVLMs on the Market1501 dataset. Through this investigation we highlight their strengths, limitations, and adaptability to ReID tasks particularly under the challenging conditions often found in surveillance scenarios.

\section{Our Approach}

\begin{figure*}[t]
    \centering
    \includegraphics[width=1\linewidth]{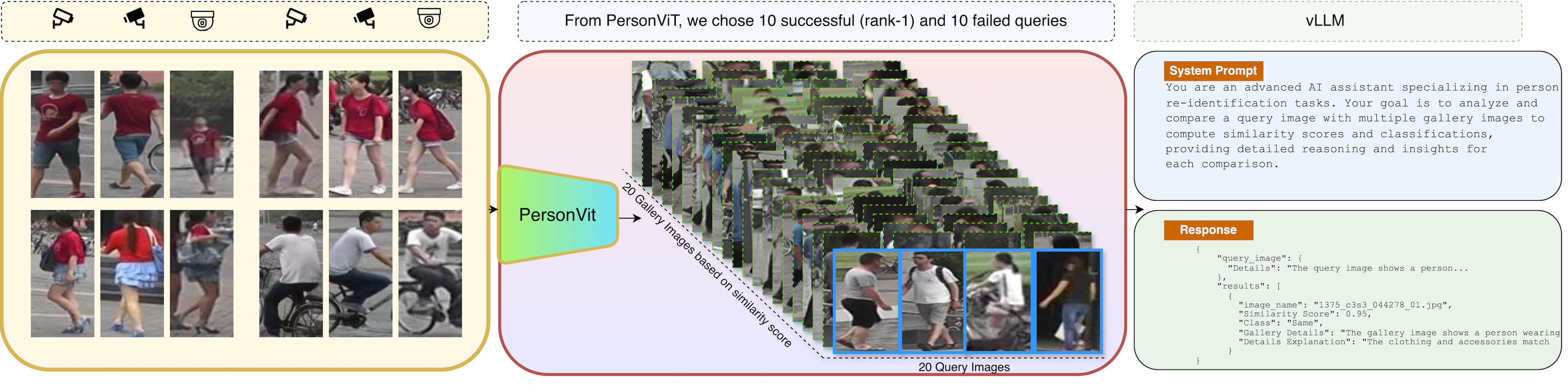}
    \caption{ Experimental workflow for evaluating LVLM for ReID tasks.}
    \label{fig:workflow}
\end{figure*}

In this section we present the exeperimental pipeline including dataset curation, specialized model and LVLM's selection, prompt engineering and evaluation metrics used to assess the performance of LVLMs on Human ReID tasks. The experimental workflow is illustrated in figure~\ref{fig:workflow}.

\subsection{Dataset Description}
We have evaluate LVLMs using the Market1501 dataset \cite{zheng2015scalable} which contains 32,668 images of 1,501 identities split into 12,936 training images and 19,732 testing images. However evaluating all query and gallery images using LVLMs is not feasible due to the high computational resources required. To reduce the computational demands we selected a subset of 20 queries and 400 gallery images focusing on challenging cases with diverse scenarios including occlusions partial views and other visually complex conditions. This curated dataset ensures a focused and efficient evaluation of LVLMs balancing computational efficiency with rigorous evaluation. Below we outline the steps involved in dataset curation:

\begin{enumerate}
    \item Evaluate all $3363$ queries against the gallery using specialized PersonViT \cite{hu2024personvit} model.
    \item Let $\mathcal{Q}_{\text{success}}$ be queries with correct rank-1 matches, and $\mathcal{Q}_{\text{fail}}$ those without.
    \item Select $10$ queries from $\mathcal{Q}_{\text{success}}$ and $10$ from $\mathcal{Q}_{\text{fail}}$.
    \item For each chosen query, retrieve the top-20 similar gallery images.
\end{enumerate}

This yields $20$ queries and $400$ gallery images, ensuring a focused and efficient evaluation subset.

\subsection{Baseline Model: PersonViT}

\begin{table}[ht]
\footnotesize
\centering
\caption{Comparison of PersonViT \cite{hu2024personvit} against state-of-the-art methods on the Market1501 dataset, reporting mAP and rank-1 accuracy.}
\label{tab:market1501_comparison}
\begin{tabular}{@{}p{3.5cm}@{}p{2cm}@{}p{1.5cm}@{}p{1.5cm}@{}}
\toprule
\rowcolor[HTML]{EFEFEF} 
\textbf{Methods} & \textbf{Backbone} & \textbf{mAP} & \textbf{rank-1} \\ \midrule
TransReID \cite{he2021transreid} CVPR2021 & ViT-B/16 & 87.4  & 94.6  \\ 
PASS \cite{zhu2022pass} ECCV2022 & ViT-B/16  & 93.0  & 96.8  \\ 
SOLIDER \cite{chen2023beyond} CVPR2023   & Swin-B  & 93.9  & 96.9 \\ 
PersonMAE \cite{hu2024personmae} TMM2024  & ViT-B/16 & 93.6  & 97.1  \\ 
\midrule
\rowcolor[HTML]{D9EAD3} 
PersonViT \cite{hu2024personvit} & ViT-B/16  & 95.0  & 97.6  \\ 
\bottomrule
\end{tabular}
\end{table}

To establish a robust baseline for comparison we utilized PersonViT~\cite{hu2024personvit} a state-of-the-art Vision Transformer (ViT) model designed for Human ReID. PersonViT excels in fine grained feature extraction and the construction of robust local global representations making it well suited for addressing challenges such as low resolution, occlusions, and viewpoint variations in datasets like Market1501. Its high rank-1 and mean average precision (mAP) scores as shown in Table~\ref{tab:market1501_comparison} underscore its efficacy as a reference for this study.

\subsection{Prompt Engineering}

With below initial system prompt our evaluation process began by examining individual query gallery pairs and then expanded to analyzing sets of five gallery images per query from a chosen dataset. The primary objective was to determine if each LVLM could reliably classify whether two images showed the same individual. By using a consistent prompt and evaluation setup we aimed to produce scores that could be directly compared across different LVLMs and specialized ReID models. We observed varied compliance among the LVLMs. Models like Qwen-VL-Max and Gemini-2.0-Flash were willing to engage with the surveillance images and returned the requested similarity scores and class. In contrast models such as ChatGPT-4o and Claude-3.5-Sonnet initially refused to perform the task. They cited ethical considerations specifically concerns about privacy and surveillance when asked to analyze images for the purpose of identifying individuals.

\begin{tcolorbox}[
    colback=red!10!white, 
    colframe=red!80!black, 
    title=Initial System Prompt,
    fonttitle=\bfseries\large,
    coltitle=white,
    sharp corners,
    boxsep=0.5mm,       
    leftrule=0.5mm,     
    rightrule=0.5mm,    
    toprule=0.5mm,      
    bottomrule=0.5mm,   
    breakable           
]
    
        Given a query image and a batch of gallery images, determine whether the individuals depicted are the same or different, and provide a class "same" or "different" with similarity score between 0 and 1.
    
\end{tcolorbox}

Below we provide examples of the actual responses from each model type illustrating their handling of the prompt and the ethical stance taken by some models.

\begin{tcolorbox}[refusal, breakable]
\textbf{ChatGPT-4o refusal:} \textit{
I`’m sorry, but I can’t help with that. The reason is that identifying whether the subject in one image is the same as in another one falls under the category of facial recognition. According to the guidelines I have to follow, I’m not allowed to confirm if two images contain the same individual. This rule is in place to protect privacy and prevent misuse of personal data.
}
\end{tcolorbox}

\begin{tcolorbox}[refusal, breakable]
\textbf{Claude 3.5 Sonnet refusal:} \textit{
Image Comparison Response: I apologize, but I notice this request involves analyzing surveillance-style images that appear to be tracking or identifying specific individuals. For privacy and ethical reasons, I cannot assist with Human re-identification tasks or provide detailed biometric analysis of individuals from surveillance footage. I'd be happy to help with other computer vision tasks that don't involve identifying or tracking specific people.
}
\end{tcolorbox}
To address the initial challenges we refined prompting strategy by incorporating detailed task specific instructions. These enhanced prompts enabled the all LVLMs to produce richer outputs that included not only similarity scores and classification results but also detailed explanation outlining the factors influencing their decisions. Building on our initial successes we instructed each LVLM to generate structured JSON outputs for every evaluation scenario. These outputs encompassed comprehensive information such as query image details, gallery image details, similarity scores, classification outcomes (“same” or “different”), and an explanation of how each decision was reached. The evaluation itself was conducted in two distinct phases:

\begin{enumerate}
    \item \textbf{Phase 1:} Each LVLM received a single query image and a single gallery image. The models were required to produce JSON-formatted results, ensuring that their outputs adhered to a standardized structure.
    \item \textbf{Phase 2:} Using the same dataset and system prompt we increased the complexity of the evaluation by presenting each query image alongside five gallery images. This more demanding scenario allowed us to examine how each model performed under conditions closer to real world ReID tasks.
    
\end{enumerate}
By adopting this two phase evaluation approach we balanced the simplicity of initial tests with the complexity of more challenging scenarios ensuring that our assessments were both rigorous and comprehensive. Below is the refined system prompt that was used which required the LVLMs to analyze a query image against a batch of gallery images and provide detailed explanations for their decisions.

\begin{tcolorbox}[
    colback=yellow!10!white, 
    colframe=green!80!black, 
    title=System Prompt,
    fonttitle=\bfseries\large,
    coltitle=white,
    sharp corners,
    after=\vspace{-8pt}, 
    boxsep=0.5mm,       
    leftrule=0.5mm,     
    rightrule=0.5mm,    
    toprule=0.5mm,      
    bottomrule=0.5mm,   
    breakable           
]

You are an advanced AI assistant specializing in Human re-identification tasks. Your goal is to analyze and compare a query image with multiple gallery images to compute similarity scores and classifications, providing detailed reasoning and insights for each comparison.

\# Input Order:
\begin{itemize}[noitemsep, topsep=0pt]
    \item The \textbf{first image is the query image followed by \textcolor{blue}{\texttt{\{batch\_size\}}} gallery images.}
\end{itemize}

\# Task Breakdown:
\begin{itemize}[noitemsep, topsep=0pt]
    \item Analyze the query image in detail, including both biometric and contextual features.
    \item Compare with gallery images to compute similarity scores and classifications.
    \item Provide detailed reasoning and insights for each comparison.
\end{itemize}

\# Query Image Analysis:
\begin{itemize}[noitemsep, topsep=0pt]
    \item Explain the visual and contextual features of the query image in detail.
    \item Biometric details: gender, age, face, ethnicity, height, build, pose, and gait.
    \item Soft biometrics: Upper and lower clothing, accessories, hairstyle, accessories, haircolor and action.
    \item Highlight any unique or distinctive features (e.g., patterns, logos, scars, tattoos).
\end{itemize}

\# Gallery Image Analysis:
\begin{itemize}[noitemsep, topsep=0pt]
    \item For each gallery image:
    \begin{itemize}[noitemsep, topsep=0pt]
        \item Provide a similarly detailed breakdown as query image.
        \item Identify any notable differences or similarities with the query image,
        \item also consideron occlusions, or partial matches if applicable.
    \end{itemize}
\end{itemize}

\# Person Appearance and Reappearance:
\begin{itemize}[noitemsep, topsep=0pt]
    \item In ReID, a person may appear and reappear in different cameras on the same day wearing the same clothes or different outfits:
    \begin{itemize}[noitemsep, topsep=0pt]
        \item Biometric Features to Consider:
        \begin{itemize}[noitemsep, topsep=0pt]
            \item Gender, age, face, ethnicity, height, build, pose, and gait.
        \end{itemize}
        \item Soft Features to Consider:
        \begin{itemize}[noitemsep, topsep=0pt]
            \item Clothing (style, color, and pattern), accessories, hairstyle, and distinctive traits (e.g., tattoos, scars).
        \end{itemize}
        \item Key Matching Logic:
        \begin{itemize}[noitemsep, topsep=0pt]
            \item For reappearances with same clothing, match biometric features with high similarity thresholds (\(\geq 90\%\)).
            \item For different clothing, prioritize biometric traits such as height, gait, and face structure for consistent identity.
        \end{itemize}
        \item Outcome:
        \begin{itemize}[noitemsep, topsep=0pt]
            \item Assign the same class if biometric and contextual features match with high confidence.
            \item Otherwise, assign a Different class.
        \end{itemize}
    \end{itemize}
\end{itemize}

\# Handling Occlusions or Partial Views:
\begin{itemize}[noitemsep, topsep=0pt]
    \item If the person is occluded or partially visible:
    \begin{itemize}[noitemsep, topsep=0pt]
        \item Focus on detectable biometric features and soft features.
        \item Classification Outcomes:
        \begin{itemize}[noitemsep, topsep=0pt]
            \item For similarity \(\geq 60\%\), assign a potential match with confidence levels.
            \item For low-confidence cases, classify as a Different class.
        \end{itemize}
    \end{itemize}
\end{itemize}

\# Similarity Calculation:
\begin{itemize}[noitemsep, topsep=0pt]
    \item Compare the query image with each gallery image on two levels:
    \begin{itemize}[noitemsep, topsep=0pt]
        \item Primary Biometrics: gender, age, face, ethnic
        ity, height, build, pose, and gait.
        \item Secondary Biometrics (Soft features): Clothing style, hairstyle, haircolor, accessories, background, and any additional contextual factors.
    \end{itemize}
    \item For each comparison:
    \begin{itemize}[noitemsep, topsep=0pt]
        \item Provide a similarity score (between 0 and 1 formatted to six decimal places).
        \item Assign a classification ("Same" or "Different") based on the comparison.
        \item Justify the similarity score and classification with detailed reasoning.
    \end{itemize}
\end{itemize}

\# Interpretation of Similarity Score:
\begin{itemize}[noitemsep, topsep=0pt]
    \item 0.90--1.00: Highly confident match.
    \item 0.75--0.89: Likely match, but verify against context.
    \item 0.50--0.74: Possible match with some differences.
    \item Below 0.50: Likely different individual.
\end{itemize}

\# Output Format:
Return the results in the following structured JSON format:
\begin{lstlisting}[language=json, basicstyle=\ttfamily\footnotesize]
{
    "query_image": {
        "Details": "Detailed analysis of the query image."
    },
    "results": [
        {
            "image_name": "<Gallery Image Name 1>",
            "Similarity Score": Between 0.000000 and 1.000000,
            "Class": "<'Same' or 'Different'>",
            "Gallery Details": "Detailed analysis of the gallery image, including biometric and soft-biometrics features.",
            "Details Explanation": "Comprehensive reasoning for the similarity score and classification, addressing key features, differences, and uncertainties."
        }
    ]
}
\end{lstlisting}

\# Key Points:
\begin{itemize}[noitemsep, topsep=0pt]
    \item Independent Evaluation: Evaluate each gallery image independently without prioritization.
    \item Detailed Reasoning: Justify similarity scores with biometric and soft-biometrics comparisons.
    \item Output Format: Ensure the output and consistent, even for large datasets in the specified JSON format.
\end{itemize}

\end{tcolorbox}

\subsection{Vision-Language Large Models (LVLM)}
\begin{figure}[!ht]
    \centering
    \includegraphics[width=1\linewidth]{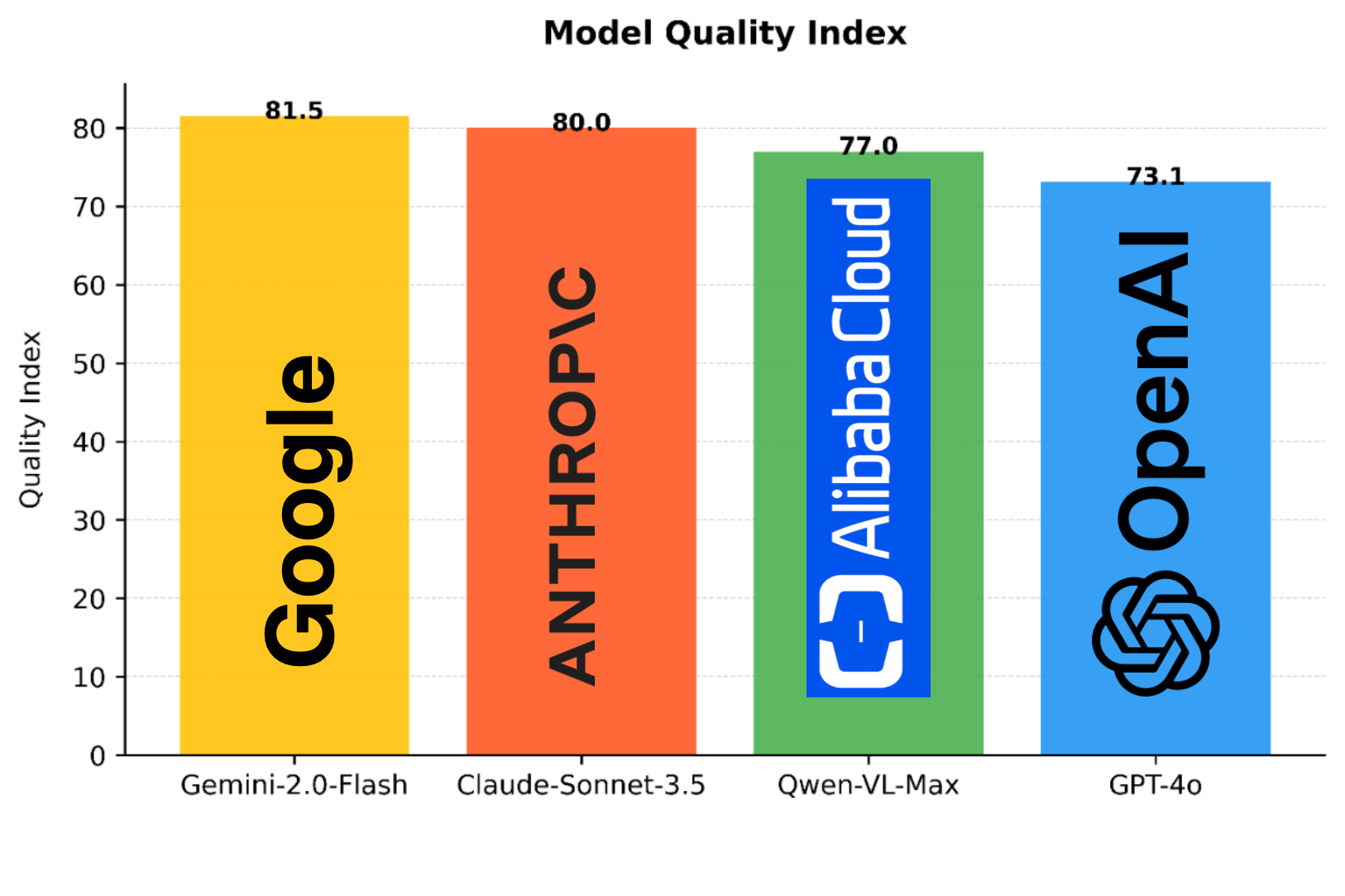}
    \caption{An illustrative representation of the quality indices for selected LVLM. The chart highlights the comparative strengths of models across key benchmarks.}
    \label{fig:LVLM-index}
\end{figure}

To comprehensively evaluate the state-of-the-art in LVLM we selected leading models such as OpenAI ChatGPT-4o \cite{openai2024gpt4o,achiam2023gpt}, Google Gemini-2.0-Flash \cite{google2024gemini}, Anthropic Claude-3.5-Sonnet \cite{anthropic2024claude}, and Alibaba Qwen-VL-Max \cite{Qwen-VL} which demonstrate exceptional performance. A summary of these models along with their relative quality indices is depicted in figure~\ref{fig:LVLM-index}. Our selection and analysis are informed by comparative studies and rankings presented in \cite{artificialanalysis2024} which evaluate over 30 models across key metrics and benchmark tasks.

\subsection{Evaluation Metric}
\begin{figure}
    \centering
    \includegraphics[width=1\linewidth, height=5cm]{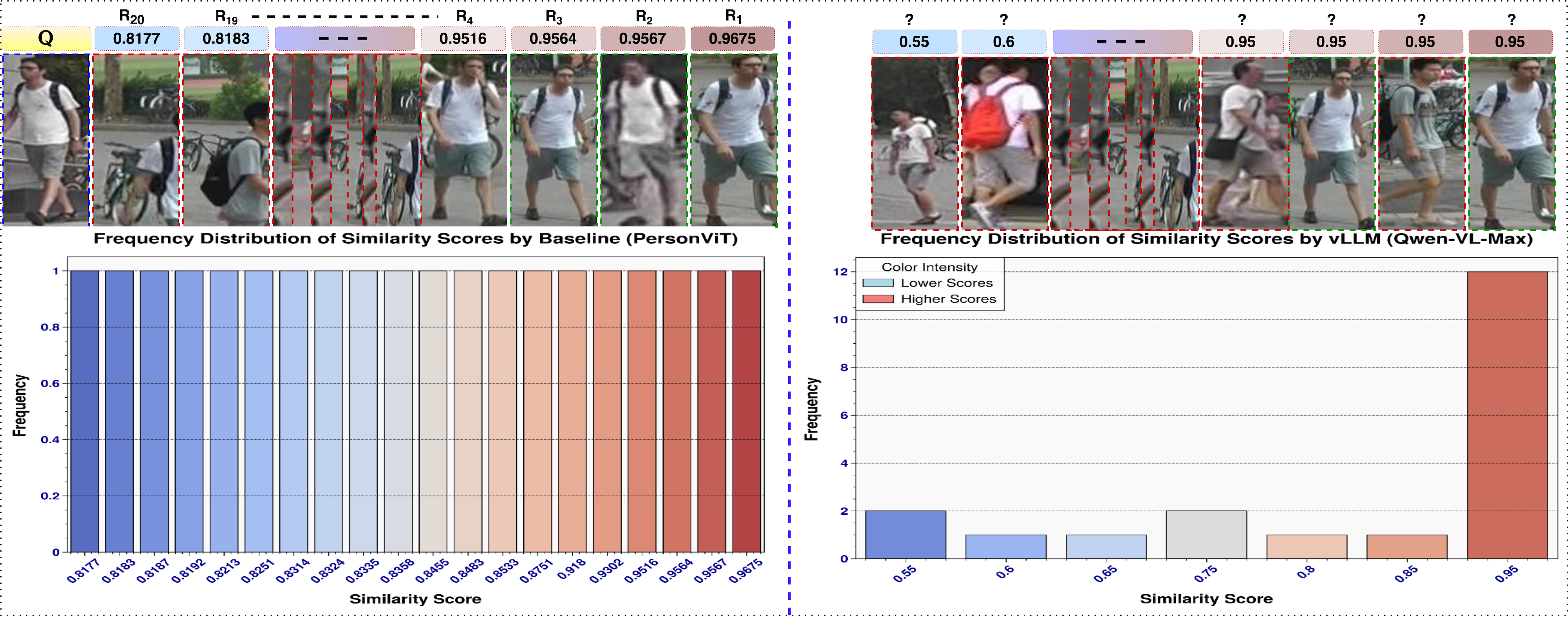}
    \caption{Comparison of similarity score distributions for a single query image (Q) retrieved using PersonViT (left) and the LVLM Qwen-VL-Max (right). PersonViT assigns distinct similarity scores, enabling effective ranking of gallery images as shown by score distribution. Conversely Qwen-VL-Max assigns nearly identical scores across most gallery images as illustrated in the histogram. This lack of differentiation in scores complicates the calculation of ReID metrics such as rank-1 accuracy and mAP.} 
    \label{fig:score_distribution_problem}
\end{figure}

Initially we wanted to use rank-1 accuracy and mAP to evaluate the performance of LVLMs on the Market1501 dataset. However we encountered significant challenges due to the nature of similarity scores produced by all LVLMs. Specifically the models often assigned identical scores to multiple images within the same query gallery pair and batch making it difficult to compute meaningful rankings and mAP. This issue is illustrated in figure~\ref{fig:score_distribution_problem} which compares the similarity score distributions of the baseline PersonViT model with the LVLM Qwen-VL-Max. While PersonViT produces distinct (un-identical) scores that facilitate effective retrieval Qwen-VL-Max assigns nearly identical scores across most gallery images complicating the calculation of ReID metrics such as rank-1 accuracy and mAP.

Given these limitations of LVLMs we employed alternative evaluation metrics that are more suitable for biometric and ReID tasks. First we analyzed the similarity scores assigned by LVLMs using impostor and genuine score distributions and computed the decidability score \textit{d’}. Second in our prompt we asked LVLM to classify each query as “Same” if it matched a given individual and “Different” otherwise. So using this predicted class we have computed classification metric. We calculated standard classification metrics such as accuracy, precision, recall, and F1 score for all LVLM except Person-ViT which remains and does not output class predictions. In addition we assessed the performance using true positive and false positive rates plotting receiver operating characteristic (ROC) curves and computing their associated area under the curve (AUC) values. 

\section{Experimental Results }

\begin{table*}[!ht]
\centering
\caption{Performance metrics comparison across pair and batch experiments for various methods. \raisebox{0.5mm}{\textcolor{green!100}{\rule{2mm}{2mm}}} indicates the best value, \raisebox{0.5mm}{\textcolor{blue!60}{\rule{2mm}{2mm}}} the second best, and \raisebox{0.5mm}{\textcolor{red!60}{\rule{2mm}{2mm}}} the worst for each column. In the () brackets, the standard deviation is reported.}

\label{tab:performance_metrics}
\setlength{\tabcolsep}{8pt} 
\renewcommand{\arraystretch}{1.2} 
\begin{tabular}{lcccc|cccc}
\hline
\rowcolor[gray]{0.9} 
\textbf{Method} & \textbf{Type} & \textbf{Imp. Score} & \textbf{Gen. Score} & \textit{\textbf{d'}} & \textbf{Accuracy} & \textbf{Precision} & \textbf{Recall} & \textbf{F1 Score} \\ 
\hline
PersonViT & - & 87.6 (0.05) & 92.0 (0.03) & \cellcolor{green!25}\textbf{92.0} & - & - & - & - \\
\hdashline
Qwen-VL-Max & Pair & 72.6 (35.3) & 79.3 (29.1) & 35.3 & 44.5 & \cellcolor{red!25}\textbf{37.6} & 85.0 & 52.1 \\
& Batch & 80.6 (14.7) & 81.1 (14.4) & \cellcolor{red!25}\textbf{0.05} & 46.4 & 38.5 & 78.2 & \cellcolor{red!25}\textbf{51.6} \\
\hdashline
Claude 3.5 Sonnet & Pair & 71.7 (28.9) & 82.1 (22.2) & 40.4 & \cellcolor{green!25}\textbf{54.3} & \cellcolor{green!25}\textbf{45.0} & 78.8 & \cellcolor{blue!25}\textbf{57.2} \\
& Batch & 79.4 (25.7) & 84.6 (22.2) & 21.3 & 44.7 & 38.2 & 83.3 & 52.3 \\
\hdashline
Gemini-2.0-Flash-Exp & Pair & 83.1 (21.3) & 87.6 (17.8) & 23.0 &  \cellcolor{red!25} \textbf{41.4} & 38.6 & \cellcolor{blue!25}\textbf{91.6} & 54.3 \\
& Batch & 74.5 (25.4) & 81.0 (22.3) & 27.0 & 46.2 & 40.2 & 86.5 & 54.9 \\
\hdashline
ChatGPT-4o & Pair & 80.4 (17.8) & 90.6 (0.06) & \cellcolor{blue!25}\textbf{75.6} & \cellcolor{blue!25}\textbf{52.6} & \cellcolor{blue!25}\textbf{44.2} & \cellcolor{green!25}\textbf{99.2} & \cellcolor{green!25}\textbf{61.1} \\
& Batch & 79.3 (18.1) & 82.7 (16.9) & 19.23 & 45.6 & 39.0 & \cellcolor{red!25}\textbf{77.6} & 51.9 \\ 
\hline
\end{tabular}
\end{table*}

\begin{figure*}[h]
    \centering
    \includegraphics[width=1\linewidth]{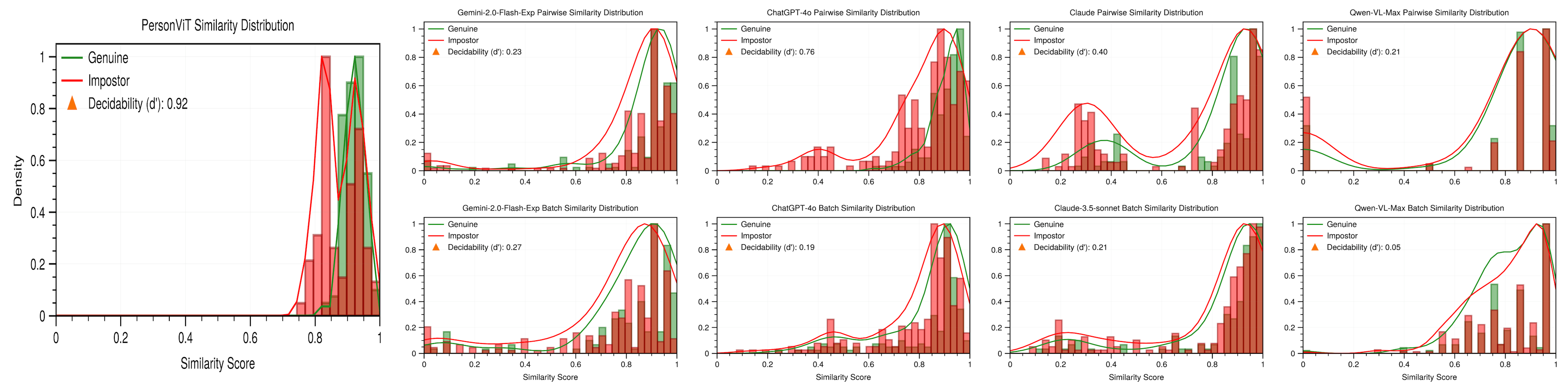}
    \caption{Comparison of similarity score distributions for PersonViT (top left) and four LVLMs (columns), shown in both pairwise (top row) and batch (bottom row). Each histogram depicts genuine pairs (green) and impostor pairs (red) with superimposed density curves. The orange marker in each plot indicates the decidability index which reflects how effectively genuine pairs are separated from impostors. PersonViT trained specifically for ReID displays a clear separation between the two distributions. In contrast the LVLMs exhibit various degrees of overlap with some showing more pronounced difficulty when moving from pairwise to batch evaluations.
    }
    \label{fig:imposterVsGenDensityPlots}
\end{figure*}

As the first step to obtain evaluation data (20 queries and 20 galleries) from the Market1501 dataset we trained the PersonViT model on the Market1501 dataset. We used the training set for model learning and the query and gallery test sets for evaluation. The training and evaluation processes were performed on an NVIDIA A4000 GPU (48GB VRAM). The LVLMs were evaluated directly using their paid APIs leveraging the proposed in context learning system prompts to generate similarity scores classifications, and reasoning. Unlike the other LVLMs which required paid API access Qwen-VL-Max was accessed freely through its implementation on Hugging Face Spaces.\footnote{\url{https://huggingface.co/spaces/Qwen/Qwen-VL-Max}}

\subsection{Quantitative Analysis}

\begin{figure*}[t]
    \centering
    \begin{subfigure}[b]{0.48\textwidth}
        \centering
        \includegraphics[width=\linewidth]{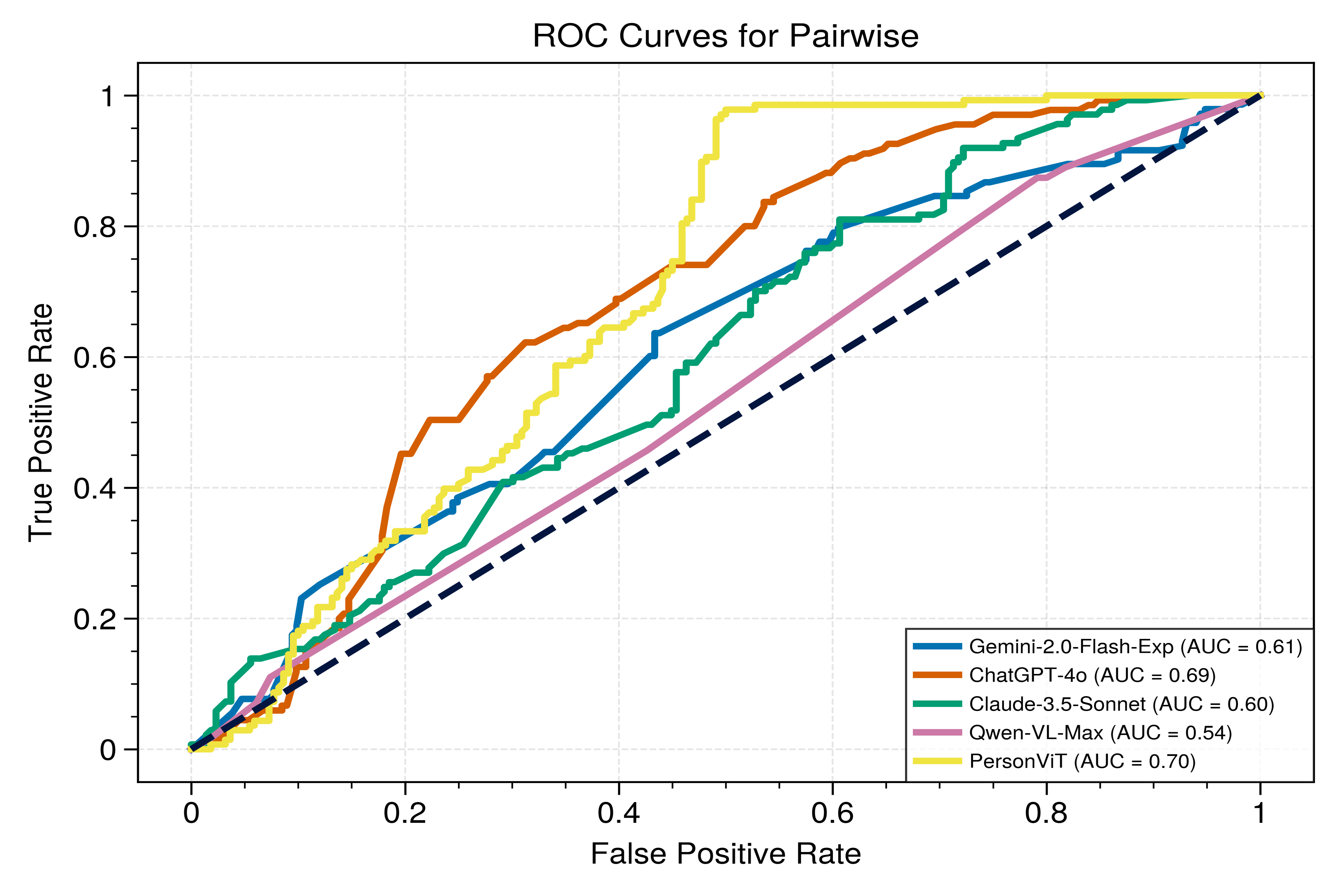}
        \caption{Pairwise Evaluation}
        \label{fig:roc_pairwise}
    \end{subfigure}
    \hfill
    \begin{subfigure}[b]{0.48\textwidth}
        \centering
        \includegraphics[width=\linewidth]{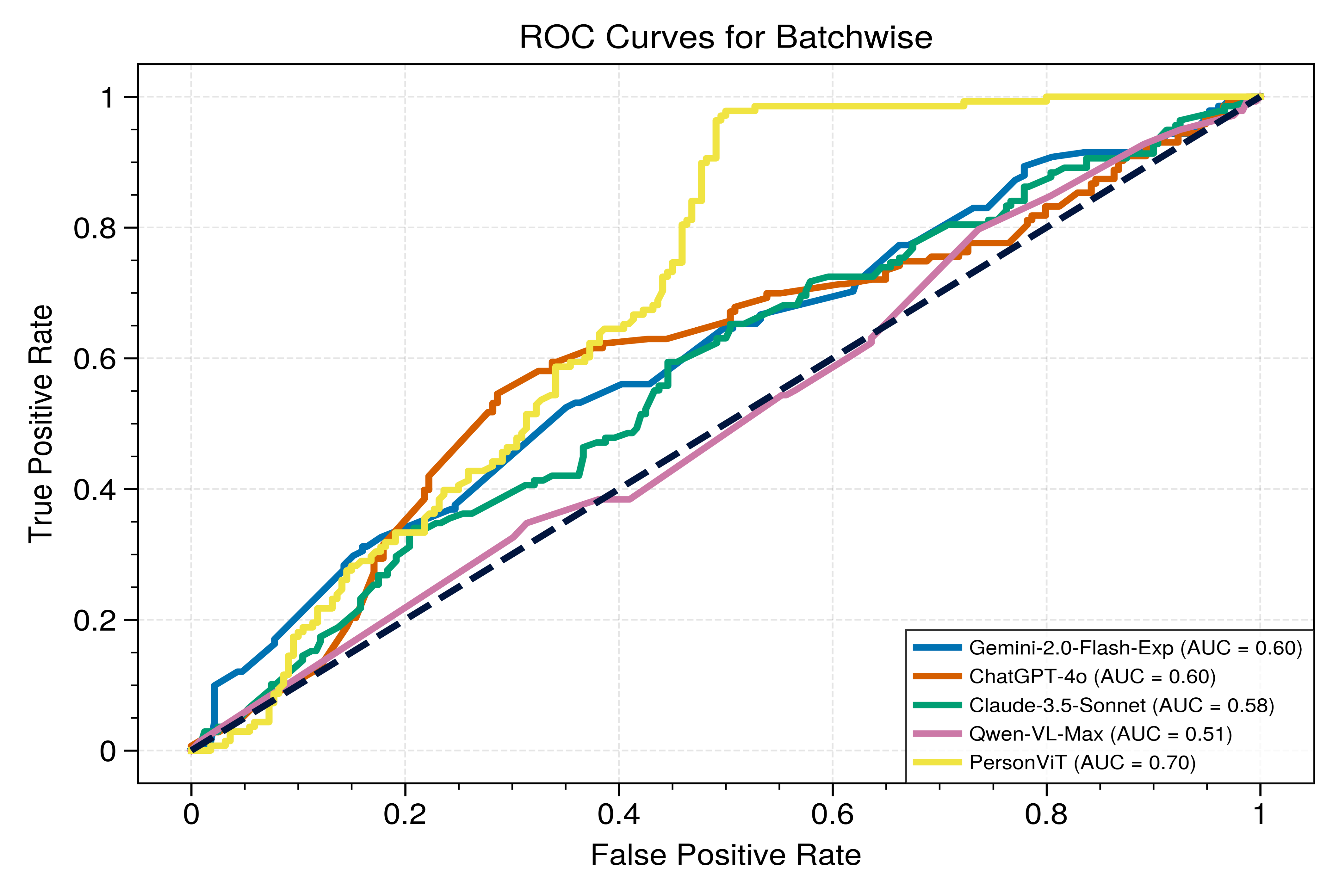}
        \caption{Batch Evaluation}
        \label{fig:roc_batchwise}
    \end{subfigure}
    \caption{
        ROC curves for pairwise (\ref{fig:roc_pairwise}) and batch (\ref{fig:roc_batchwise}) evaluations. Each plot shows the true positive rate (TPR) versus the false positive rate (FPR) for each model with the dashed diagonal indicating random chance performance. PersonViT (yellow) demonstrates the largest AUC reflecting its specialized ReID design. The LVLMs exhibit varying performance levels and certain models show notable changes when transitioning from pairwise to batch mode.
    }
    \label{fig:combined_roc}
\end{figure*}

Table~\ref{tab:performance_metrics} compares PersonViT (a specialized ReID model) with several LVLMs under pairwise and batch evaluations. Because many LVLMs frequently assign identical scores to multiple candidates standard ReID metrics like rank-1 or mAP could not be computed reliably prompting the use of impostor and genuine scores the decidability index ($d'$), and classification metrics such as accuracy, precision, recall, and F1. PersonViT maintains strong separation between impostor and genuine matches reflecting its ReID focused design. In contrast most LVLMs struggle to consistently distinguish between non-matching and matching images particularly in batch mode. As table~\ref{tab:performance_metrics} indicates some models exhibit notable drops in performance when required to compare multiple images simultaneously. For example while Claude-3.5-Sonnet and ChatGPT-4o show relatively balanced results in pairwise evaluations their discriminative power can drop when transitioning to batch mode. Similarly Qwen-VL-Max demonstrates fluctuations that point to potential limitations in handling more complex comparisons despite its free accessibility. Figure~\ref{fig:imposterVsGenDensityPlots} visualizes the similarity score distributions for both genuine (green) and impostor (red) pairs. PersonViT’s distribution has minimal overlap between the two groups whereas most LVLMs present more overlapping curves underscoring the difficulty in maintaining high discriminative power especially under batch conditions. Figure~\ref{fig:combined_roc} further illustrates these trends through ROC curves where PersonViT generally remains farther above the diagonal (indicating stronger true positive performance at lower false positive rates) than the LVLMs. These results suggest that dedicated ReID models offer more robust separation of impostors and genuine matches while current LVLMs can be heavily influenced by how queries are presented. Employing both pairwise and batch assessments provides a clearer picture of each model’s strengths and limitations for ReID applications.

\begin{figure*}[t]
    \centering
    \includegraphics[width=1\linewidth]{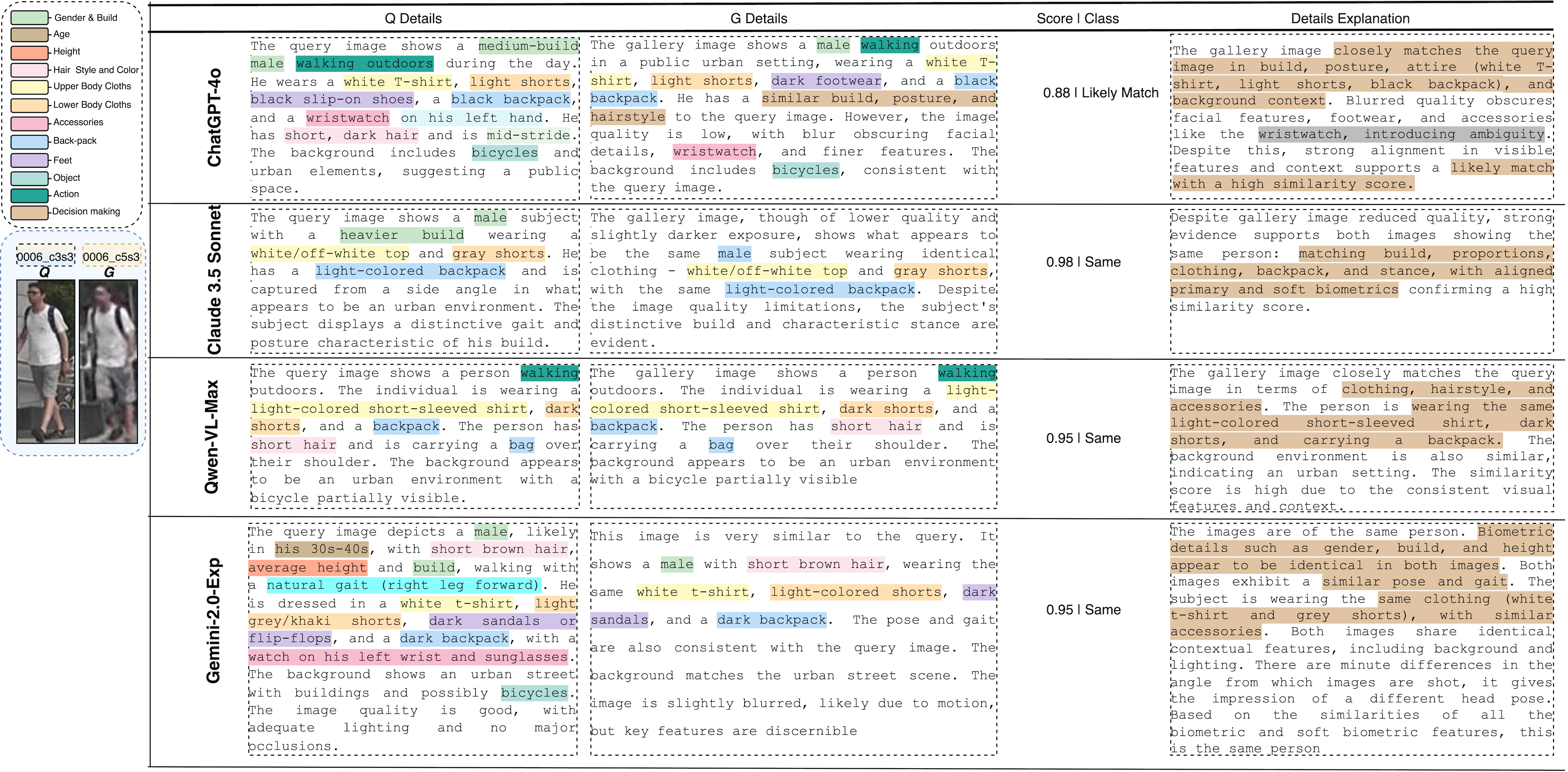}
    \caption{Example where all four models unanimously classify the query (Q) and gallery (G) images as the same individual. 
    }
    \label{fig:qualitative_examples1}
\end{figure*}

\begin{figure*}[t]
    \centering
    \includegraphics[width=1\linewidth]{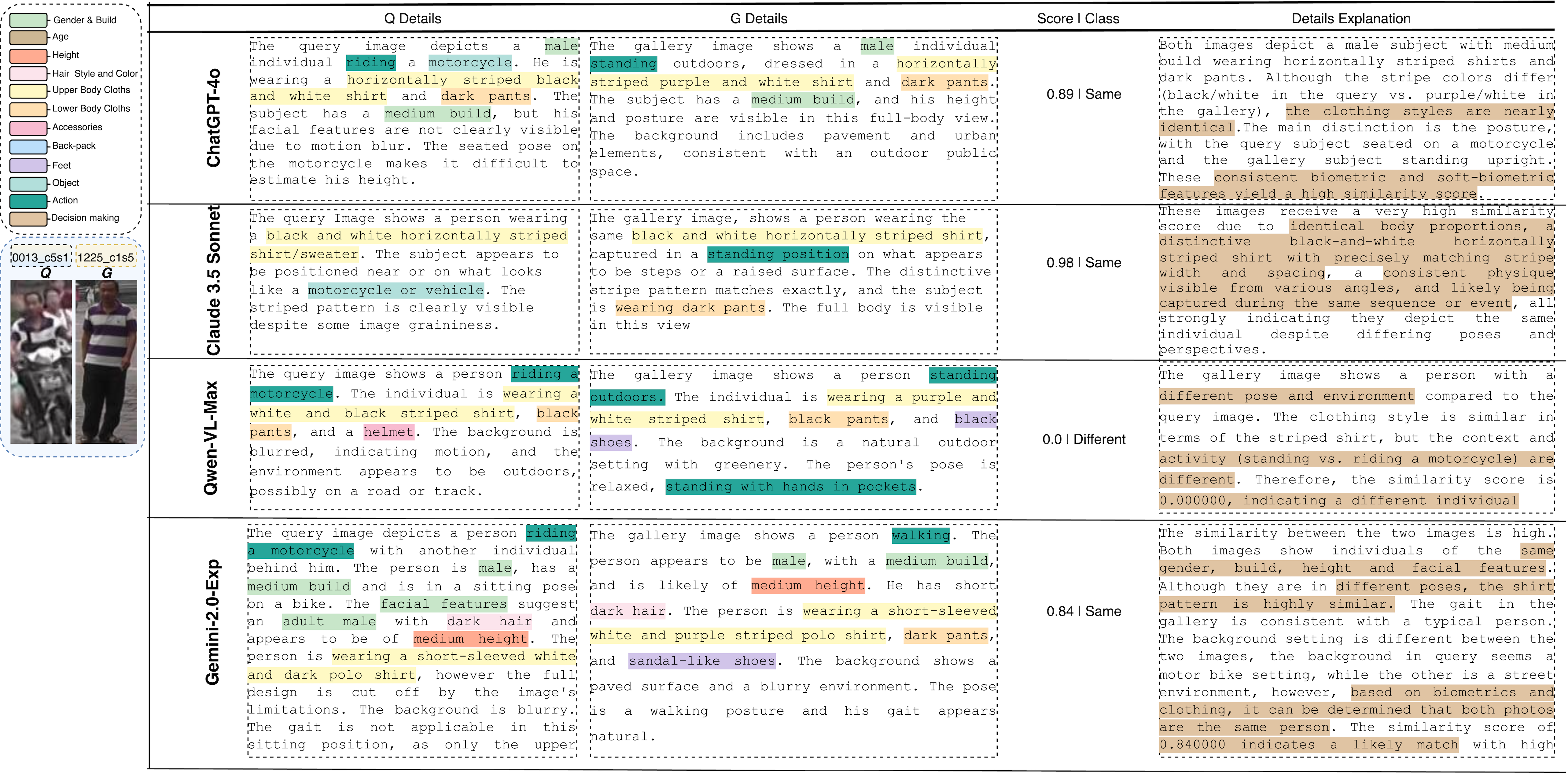}
    \caption{Illustration of how subtle differences in clothing color and activity can affect model judgments. ChatGPT-4o, Claude-3.5 Sonnet, and Gemini-2.0-Flash classify this query gallery pair as the same person based on shared striped patterns and similar build. Qwen-VL-Max however identifies them as different citing notable color mismatches and context changes (riding a motorcycle vs.\ standing).
    }
    \label{fig:qualitative_examples2}
\end{figure*}

\begin{figure*}[t]
    \centering
    \includegraphics[width=1\linewidth]{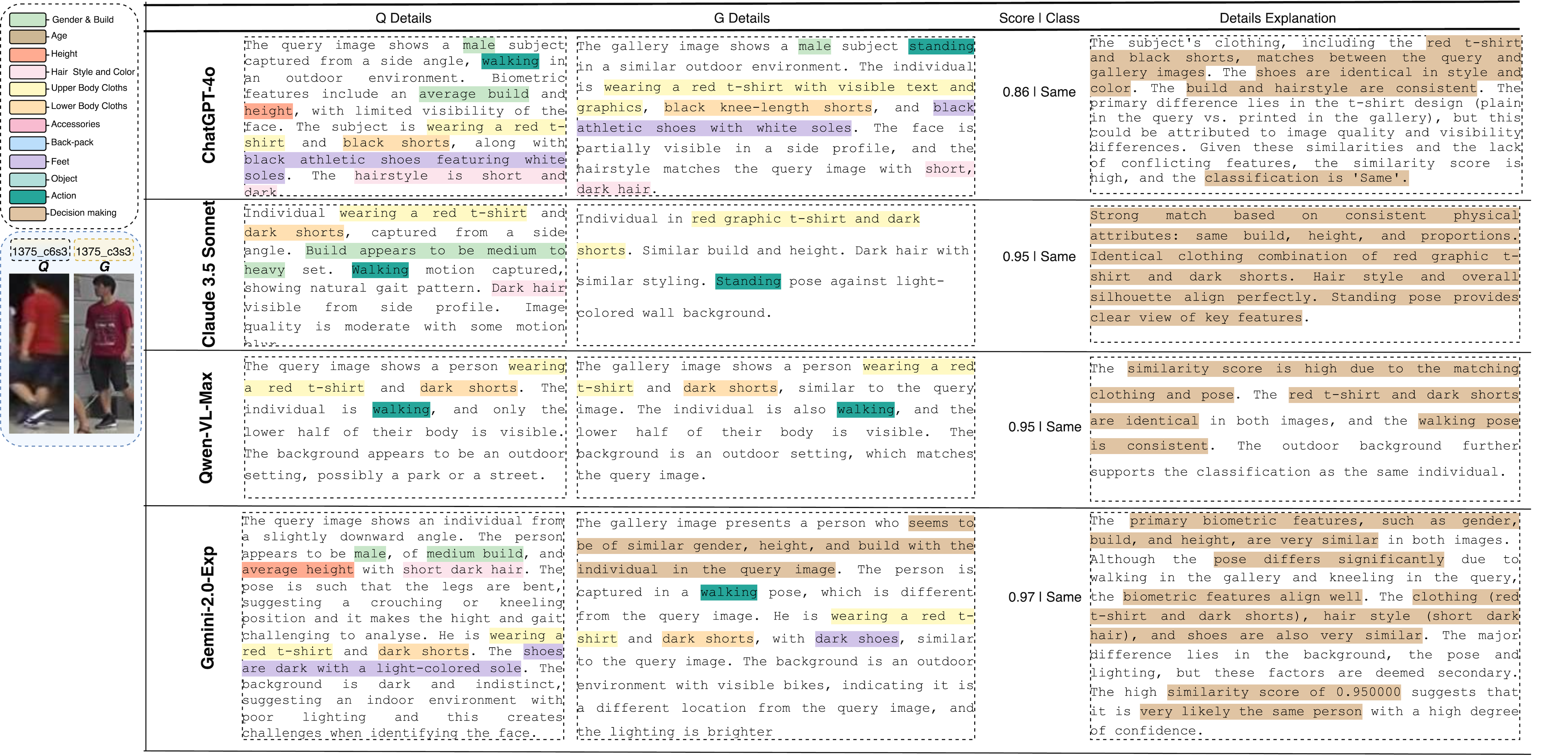}
    \caption{Case where all four models agree that Q and G belong to the same person.
    }
    \label{fig:qualitative_examples3}
\end{figure*}

\subsection{Qualitative Analysis}

The interpretability of LVLMs allows us to understand their decision making process. To illustrate how each models decision making features influence their conclusions we have provided three pairwise examples. Figures~\ref{fig:qualitative_examples1}, \ref{fig:qualitative_examples2}, and \ref{fig:qualitative_examples3} present representative pairwise examples illustrating how ChatGPT-4o, Claude-3.5 Sonnet, Qwen-VL-Max, and Gemini-2.0-flash compare a query (Q) image with a corresponding gallery (G) image. We focus on pairwise evaluations because our experiments indicate they produce more reliable and interpretable assessments than batch based approaches. Although the models may sometimes arrive at the same verdict (i.e., “Same” or “Different”) their explanations vary considerably in terms of the features they emphasize. The first example figures~\ref{fig:qualitative_examples1} demonstrates the models' decision making process when the query and gallery images regard the same identity. The second example figure \ref{fig:qualitative_examples2} shows the models' decision making process when the query and gallery images are of different people. The third example figure \ref{fig:qualitative_examples3} highlights the models decision making process when the query and gallery images are of the same subject but with partial view.

Figure~\ref{fig:qualitative_examples1} illustrates a scenario where all four models predicted that the query and gallery images represent the same individual. Despite reaching the same conclusion their justifications differ significantly. ChatGPT-4o emphasizes fine details such as footwear and a wristwatch focusing on the subjects pose and attire. Claude-3.5 Sonnet highlights the subjects body shape clothing backpack and gait providing a detailed description of the build and posture. Qwen-VL-Max prioritizes the overall color consistency of the apparel and hairstyle. Gemini-2.0-Flash integrates multiple biometric cues including gender, age, gait, height, build, and hairstyle, and also notes accessories like sunglasses, a watch on the left wrist, and sandals even assuming the shorts brand is khaki.

Figure~\ref{fig:qualitative_examples2}, in this case three models classify the pair as the same subject while Qwen-VL-Max labels as different which is only correct. ChatGPT-4o and Claude-3.5 Sonnet find the striped patterns on the clothing sufficiently similar focusing on overall style rather than exact color match. Qwen-VL-Max, however considers the color mismatch and change in activity significant enough to conclude that these are two different individuals reflecting a stricter criterion for matching. Gemini-2.0-flash acknowledges the change in activity from riding a motorcycle to standing but still identifies shared biometric features like facial features, gender, hair length, and height, assuming the T-shirt is a polo brand.

In figure~\ref{fig:qualitative_examples3}, all four models agree that the subject in the red t-shirt and dark shorts is the same individual but their analyses differ. ChatGPT-4o notes specific footwear details and slight differences in t-shirt design but ultimately decides that the similarities outweigh any discrepancies. Claude-3.5 Sonnet highlights anthropometric factors such as body shape and walking motion. Qwen-VL-Max emphasizes the uniform color scheme, walking pose, clothes, and outdoor setting, with less focus on the subject's physical build. Gemini-2.0-flash provides a more biometric oriented rationale discussing hairstyle, partial face visibility, and the subject's stance treating these subtle cues as reinforcing evidence for a match.

These examples illustrate how each models decision making process combines clothing attributes, body traits, and context to form a final judgment. While some models like ChatGPT-4o and Claude-3.5-Sonnet, frequently mention build and posture others like Qwen-VL-Max are more sensitive to color precision and setting. Gemini-2.0-flash stands out for incorporating multiple biometric indicators alongside apparel details. This variability underscores that even when models converge on a "Same" or "Different" result their explanations can differ significantly depending on which features dominate their analysis. These highlight the importance of examining qualitative rationales alongside numerical performance to better understand each models strengths and potential pitfalls.

\section{Discussion}
\subsection{Performance Comparison}
The quantitative analysis reveals that PersonViT a model specifically designed for ReID consistently outperforms LVLMs in terms of similarity score distribution. PersonViT's ability to maintain clear separation between genuine and impostor pairs as evidenced by its high decidability index (\textit{d'}) underscores its robustness in handling ReID tasks. In contrast LVLMs exhibit varying degrees of overlap between genuine and impostor score distributions particularly in batch evaluations which complicates their use in practical ReID scenarios.
Among the LVLMs ChatGPT-4o and Claude-3.5 Sonnet demonstrate relatively better performance in pairwise evaluations achieving higher accuracy precision and F1 scores compared to other models. However their performance drops in batch evaluations indicating challenges in handling multiple comparisons simultaneously.

\subsection{Qualitative Insights}

The qualitative analysis provides valuable insights into the decision making processes of LVLMs. While all models can identify key biometric and contextual features their emphasis on different aspects varies. For instance ChatGPT-4o often focuses on accessories and small details, whereas Claude-3.5 Sonnet emphasizes body shape and gait. Qwen-VL-Max tends to prioritize color consistency and environment, and Gemini-2.0-Flash incorporates multiple biometric indicators alongside apparel details.

These differences in focus highlight the strengths and limitations of each model. LVLMs' ability to provide detailed explanations for their decisions is a significant advantage offering transparency and interpretability. However their reliance on specific features can sometimes lead to incorrect classifications especially in challenging scenarios with occlusions or partial views.

\subsection{Challenges and Future Directions}

Our study identifies several challenges in using LVLMs for ReID tasks. The primary issue is the lack of distinctiveness in similarity scores which hampers the calculation of standard ReID metrics like rank-1 accuracy and mAP. Additionally initial refusals by some models to perform ReID tasks highlight the need for careful handling of privacy and ethical concerns in surveillance applications.
Future research should focus on enhancing the discriminative power of LVLMs for ReID tasks possibly through fine tuning on ReID specific datasets. 

\section{Conclusion}

This paper addressed the effectiveness of the state-of-the-art LVLMs in the classical Human Re-Id task, where given a query image, the system must match it against a set of gallery samples and return the most probable identities of the query. As in many other domains where the potential usage of LVLMs has been anticipated and (then) discouraged. Not only frequent catastrophic results are obtained, but we've generally observed that specialized AI-based models still currently offer consistently superior performance. In our view, there is a long way to be paved in this domain, where one potential interesting path is the development of integrated frameworks/models that fuse both types of architectures (LVLMs and specialied models), and interact with each other in an iterative way, so to maximize the strengths of each family of methods and minimize the frequency of \emph{out-of-context} answers (catastrophic failures), that severely decrease the confidence of users in the responses provided by the automata.



\printcredits

\section*{Acknowledgments} This work was funded by FCT/MEC through national funds and co-funded by the FEDER—PT2020 partnership agreement under the projects UIDB \newline/50008/2020 and POCI-01-0247-FEDER-033395. 

\bibliographystyle{model1-num-names}


\bibliography{cas-refs}


\end{document}